\documentclass[11pt]{article}

\usepackage[preprint]{acl}

\usepackage{amsmath,amssymb,amsfonts}
\usepackage{graphicx}
\usepackage{booktabs}
\usepackage{array}
\usepackage{xcolor}
\usepackage{url}
\usepackage{pifont}   
\usepackage{colortbl} 
\usepackage{tikz}
\usetikzlibrary{arrows.meta,positioning,fit,backgrounds,calc}

\usepackage{times}
\usepackage{latexsym}

\usepackage[T1]{fontenc}

\usepackage[utf8]{inputenc}

\usepackage{microtype}

\usepackage{inconsolata}

\usepackage{graphicx}

\newcommand{\method}{\textsc{CLAM}}
\newcommand{\ttt}{\textcolor{orange!85!red}{\ding{72}}}
\IfFileExists{fontawesome5.sty}{%
  \usepackage{fontawesome5}%
  \renewcommand{\ttt}{\textcolor{orange!85!red}{\faFire}}%
}{}
\newcommand{\dw}{\Delta W}
\newcommand{\Sset}{\mathcal{S}}

\newcommand{\tbest}[1]{\cellcolor{blue!22}\textbf{#1}} 
\newcommand{\second}[1]{\cellcolor{blue!8}#1}          

\title{\protect\raisebox{-0.22\height}{\includegraphics[height=2em]{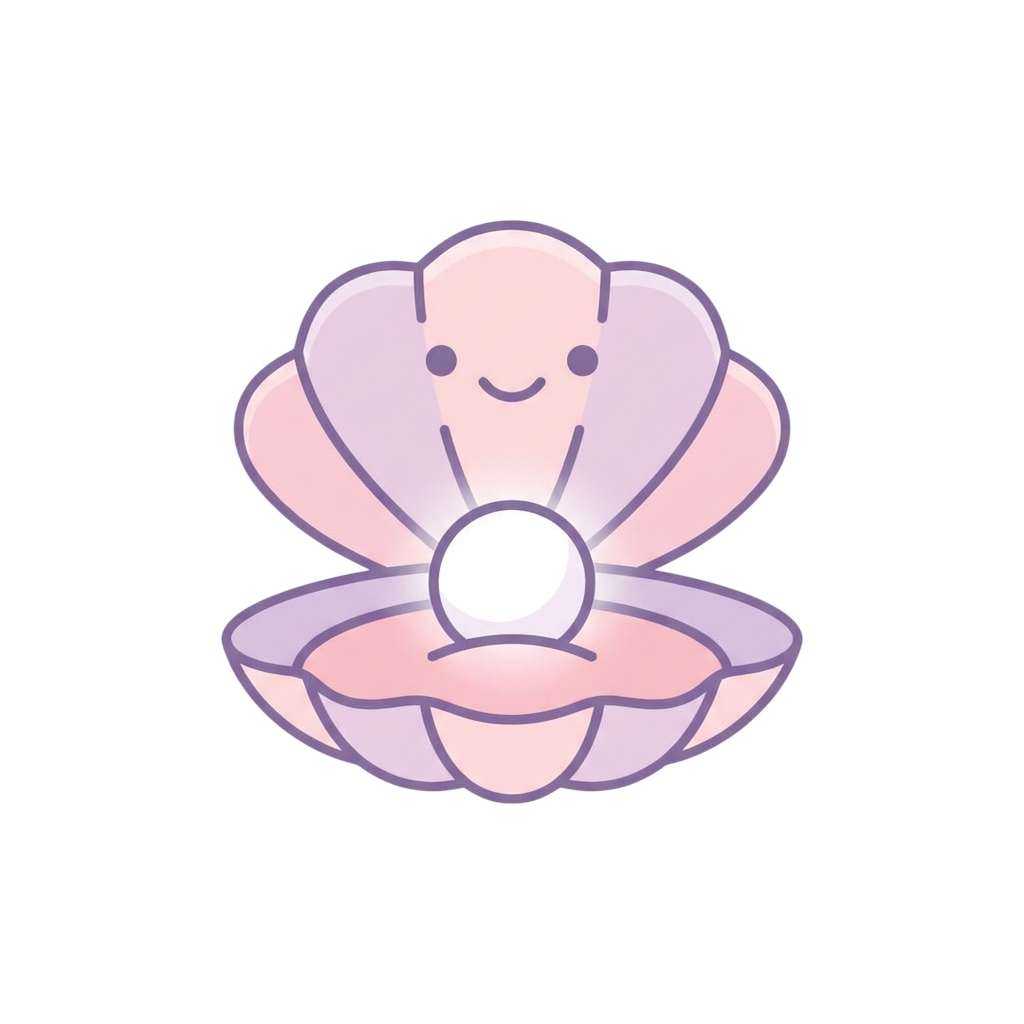}}\hspace{0.4em}Instant Weight-Space Consolidation for Long-Term Memory in LLMs}

\author{Manan Suri, Dinesh Manocha \\
  University of Maryland, College Park \\
  \texttt{\{manans, dmanocha\}@umd.edu} \\}

\begin{document}

\maketitle

\begin{abstract}
Agentic memory systems give LLMs persistent memory by extracting experience into
text and retrieving it at query time, but the model itself doesn't evolve:
however many interactions it has had, it starts each query the same. Extraction
errors compound, implicit cross-session patterns that no single fragment states
cannot be surfaced, and the construction cost grows with every conversation.
Fine-tuning on past interactions does change the weights, but requires
backpropagation at serving time and degrades on long histories. We introduce
\method\ (Carried Latent Adapter Memory), which combines the strengths of both: a
recurrent hypernetwork performs \emph{parametric consolidation}, compiling a stream
of session activations into a low-rank adapter in a single forward pass with no
gradient steps and no language-model generation calls, while a complementary
\emph{episodic buffer} supplies verbatim detail at inference. A gated latent state
carries knowledge across sessions in fixed size, so the parametric memory accumulates without
growing. On LoCoMo and LongMemEval-S across three open models (2--7\,B
parameters), \method\ leads all agentic baselines by 1--45 points and all
gradient-based parametric methods by 10--45 points, while requiring zero generative LLM calls,
zero gradient steps, and 3--151$\times$ less wall-clock time than extraction systems
to build its memory.  The two channels are complementary: the consolidated adapter adds 12--15 accuracy
points when the episodic buffer misses the gold evidence, and still adds 4--5
points even with an oracle buffer.
\end{abstract}

\section{Introduction}

Biological memory is not a database that an organism queries; it is a multi-level
process in which two complementary systems cooperate. A fast episodic store, seated
in the hippocampus, captures specific experiences rapidly and without interference.
A slow semantic store, residing in neocortical synapses, integrates those episodes
over time into structured, persistent knowledge
\cite{tulving1972episodic,squire1992memory}. The bridge between them is
\emph{consolidation}: transient experience becomes a durable change in synaptic
weights, and after consolidation the organism is physically different.
Complementary Learning Systems (CLS) theory formalizes why both stores are
necessary: the fast store gives exact recall of specific episodes without
overwriting prior knowledge, the slow store gives generalization and cross-episode
integration, and neither alone suffices
\cite{mcclelland1995complementary,kumaran2016complementary}. They are complementary
precisely because they trade off different failure modes: the fast store is lossless
but unintegrated, the slow store is integrated but lossy. A system that has only
one operates with a fundamental blind spot.

Contemporary LLM memory systems are sophisticated episodic stores. Agentic
frameworks extract facts and summaries from conversations, organize them as
profiles or knowledge graphs, and retrieve relevant fragments into the context
window at query time
\cite{chhikara2025mem0,rasmussen2025zep,packer2023memgpt,xu2025amem,park2023generative,zhong2024memorybank,simplemem2026}.
This is effective when the needed information was explicitly stated, correctly
extracted, and successfully retrieved, but each step introduces error that
compounds across the pipeline: extraction misses implicit facts, indexing loses
context, and retrieval surfaces the wrong fragment for the query at hand.
More fundamentally, the model's weights never change. Implicit patterns that no
single fragment states, cross-session structure that requires integration, and
preferences that accumulate gradually over many interactions have no pathway into
the model. In CLS terms, these systems have a sophisticated hippocampus and no
neocortex: no consolidation pathway turns experience into weight-resident knowledge.

The natural remedy is to write experience into the weights directly. A nascent line
of work does so by fine-tuning at test time: generating QA pairs from the
conversation and training on them with SGD \cite{tmem2026}, or adapting on
unlabelled context with an entropy gate \cite{hu2025tlm}. This does produce a
parametric change, but it requires backpropagation at serving time, which is
operationally costly and slow. More fundamentally, as our experiments show, gradient
based consolidation degrades on long histories: the gradient signal becomes noisy,
the model overfits early sessions, and performance drops as the history grows. It
is not the stable, accumulating consolidation that long-term memory requires.

We take a different approach that combines the strengths of both paradigms. A
hypernetwork, trained once offline, can map session activations to adapter weights
in a single forward pass: the persistence of a weight change at the speed of
inference, with no gradient step at test time. We build \method\ (Carried Latent
Adapter Memory) on this observation. A recurrent hypernetwork performs
\emph{parametric consolidation}: it ingests each session's activations from the
frozen base LLM and folds them into a compact latent state through a learned, gated
recurrence; a single forward pass over the accumulated state emits the low-rank
adapter the model wears while answering. A complementary \emph{episodic buffer}
supplies verbatim detail at inference for what a bounded parametric state cannot
retain with perfect fidelity. The parametric consolidation channel plays the role of
slow, integrated knowledge; the episodic buffer plays the role of fast, lossless
recall; and we find they are complementary.

\paragraph{Main results.}
On LoCoMo and LongMemEval-S across three open base models (Gemma-2-2B
\cite{gemmateam2024gemma2}, Qwen3-4B \cite{qwen2025qwen3}, Mistral-7B
\cite{jiang2023mistral}), \method\ leads all agentic baselines by 1--45 points and all
gradient-based parametric methods by 10--45 points, with 95\% bootstrap confidence
intervals that exclude zero in 28 of 30 paired comparisons. It does this with zero generative
language-model calls, zero gradient steps, and zero generated tokens to build its
memory: one hypernetwork forward pass per conversation, 3--151$\times$ faster than
extraction systems that spend 12--464 generative LLM calls and $3{\times}10^4$ to
$6.2{\times}10^5$ generated tokens on the same conversation, and 1.3--16$\times$
faster than gradient-based methods that take 18--63 backward passes. The two
channels validate the complementary-systems design: adding parametric consolidation
on top of the same episodic buffer is worth +7 to +11 points overall; the adapter
contributes +12 to +15 points on the 36\% of questions where the buffer fails to
surface the gold evidence; and it still adds +4 to +5 points even when the buffer
is given oracle access to the gold spans, confirming that the consolidated weights
carry information the episodic channel cannot supply rather than a better-indexed
copy of it. \textbf{Contributions.}
\begin{itemize}\itemsep2pt
\item \textbf{\method}: a recurrent hypernetwork that performs parametric
consolidation of interaction sessions into a low-rank adapter in a single forward
pass with a carried latent state, paired with an episodic buffer in a dual-channel
architecture inspired by complementary learning systems.
\item \textbf{A scalable multi-session data pipeline}, used to generate up to
$\sim$900K high-quality, memory-dependent training trajectories.
\item \textbf{Rigorous evaluation} across two benchmarks, three model scales, and
twelve baselines spanning both agentic and parametric paradigms.
\end{itemize}


\begin{figure*}[h]
\centering
\includegraphics[width=\textwidth]{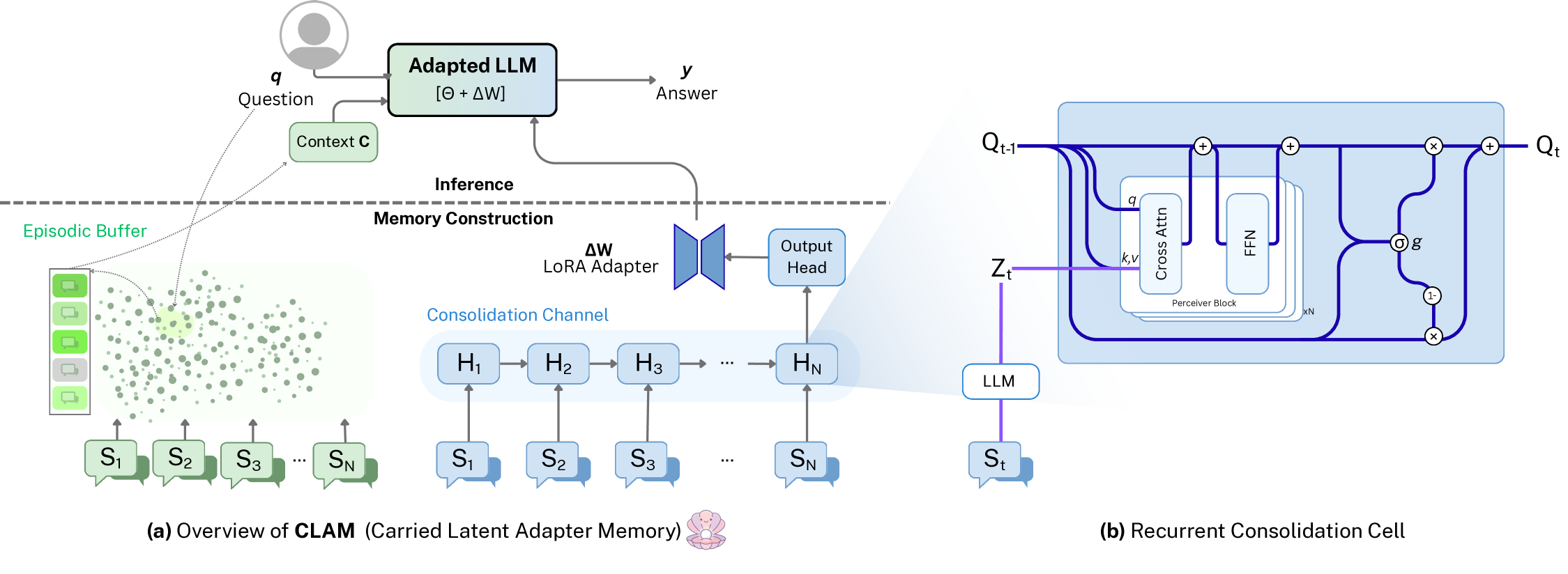}
\caption{\textbf{(a)}~\method\ overview: sessions are consolidated into a carried
workspace $Q_t$ via a recurrent hypernetwork that decodes $Q_N$ into a LoRA adapter;
the episodic buffer retrieves spans at query time.
\textbf{(b)}~One consolidation step: $Q_{t-1}$ cross-attends into session
activations $Z_t$, slots reorganise via self-attention, and a gate controls the
write.}
\label{fig:arch}
\end{figure*}
\section{Related Work}

\subsection{Memory Systems for LLMs}
The dominant paradigm stores interaction history externally and retrieves it at
query time. \textbf{Flat retrieval} indexes raw conversation chunks and prepends
the top matches \cite{lewis2020rag,guu2020realm}.
\textbf{Profile-plus-episodic} systems additionally extract structured user
facts \cite{chhikara2025mem0}. \textbf{Graph-based} systems organize memory as a
(temporal) knowledge graph for multi-hop traversal \cite{rasmussen2025zep,gutierrez2024hipporag}.
\textbf{Agentic} architectures manage memory with explicit read/write tools:
MemGPT/Letta treats context like paged virtual memory \cite{packer2023memgpt},
and Generative Agents periodically reflect episodic observations into higher-level
summaries \cite{park2023generative}. Others add forgetting dynamics \cite{zhong2024memorybank}, compression \cite{fang2025lightmem}, self-organizing links \cite{xu2025amem},
or intent-aware retrieval over compressed memory units \cite{simplemem2026}.

Despite their diversity, these systems share one property: \emph{the base model's
parameters never change}. Whatever consolidation they perform, summarizing,
re-indexing, graph-building, happens in token space and is re-read as context.
They are thus non-parametric; most require per-item LLM extraction, so memory is
not built instantly; none accumulate in a \emph{bounded} parametric state that improves with
more sessions (their stores grow without bound); and knowledge is not persistent
across queries without re-reading history. \method\ is parametric, instant,
accumulating, and persistent at once, which motivates our design. A nascent line
does modify weights to consolidate memory over time \cite{behrouz2026sleep}; \method\
shares this goal but writes memory through a single hypernetwork forward pass rather
than gradient-based self-modification, and pairs it with episodic recall.

\subsection{Hypernetworks and Context Distillation}
\method\ belongs to a growing line of work that \emph{generates} adapter weights
with a hypernetwork \cite{ha2017hypernetworks}, paired with
context distillation, which trains a model to behave as if a context were present
without reading it \cite{snell2022learning,askell2021general};
recent work distills context into a latent memory rather than into weights
\cite{zheng2026contextdistillation}. Adapter
generators map an input to a low-rank adapter in one forward pass: from a document
\cite{charakorn2026doctolora}, from a task description \cite{charakorn2025texttolora}, an evolving code repository through a recurrent state \cite{hotsko2026code2lora}, a reusable skill \cite{zhao2026parametricskills},
or video frames \cite{frames2lora2026}; from an LLM used directly as a weight generator
\cite{liu2026shine}. \method\ reuses this document-to-adapter foundation but differs
on three axes. \emph{Input:} these methods condition on a single static artifact
(one document, skill, or clip); \method\ consumes a \emph{stream} of
interaction sessions. \emph{Accumulation:} to handle many inputs, most prior work
\emph{composes} one adapter per chunk by rank-concatenation or merging, so the
represented set grows with the number of chunks and the merged signal dilutes
(roughly $1/\sqrt{N}$), which does not scale. Code2LoRA-Evo \cite{hotsko2026code2lora} instead carries a recurrent state across code diffs. \method\ instead keeps a \emph{fixed}
rank and accumulates through a recurrent neural update of a fixed-size latent state,
so adapter cost and capacity stay constant as sessions grow. \emph{Setting:} none target
long-term LLM memory or pair adapter generation with a complementary
episodic-retrieval channel. We also relate to test-time training \cite{sun2020testtime,hardt2024testtime}, which adapts weights per input via gradient steps; our adaptation is a single
forward pass with no test-time optimization.


\section{Carried Latent Adapter Memory}
A user interacts with an assistant over a stream of sessions
$\Sset = (S_1, \dots, S_N)$; at query time the system must answer using knowledge
distributed across the entire stream. \method\ answers through two complementary
channels:
\begin{equation}
  y \;=\; f_{\theta + \dw}\bigl(q,\; C\bigr),
  \label{eq:overview}
\end{equation}
where $\theta$ are the frozen base-model weights;
$\dw = H_\phi(S_{1:N})$ is a low-rank adapter produced by the
\emph{consolidation channel}---a recurrent hypernetwork $H_\phi$ that compiles the
full session stream into weight-resident knowledge in a single forward pass; and
$C = \mathrm{Buffer}(q, S_{1:N})$ is local context supplied by the
\emph{episodic buffer} at query time. The consolidation channel is computed once per stream and cached; it does not depend
on $q$. The episodic buffer is per query and does not modify $\theta$.

\subsection{Consolidation Channel}
\label{sec:consolidation}

The consolidation channel maps a session stream to an adapter through three stages:
\emph{read} each session into a fixed-size latent workspace, \emph{carry} the
workspace across sessions via a gated recurrence, and \emph{decode} the final
workspace into adapter weights \cite{hu2022lora}.

\paragraph{Read.}
To encode session $S_t$, we pass it through the frozen base model and collect
per-layer hidden states $Z_t \in \mathbb{R}^{L \times |S_t| \times d}$, where $L$
is the number of transformer blocks and $d$ the model dimension. A stack of
Perceiver cross-attention blocks~\cite{jaegle2021perceiver} compresses $Z_t$ into a
fixed-size workspace: $m$ latent queries cross-attend into $Z_t$, with each slot
pooling the session independently of the others.
\paragraph{Carry.}
Memory requires the workspace to persist and accumulate. We maintain a carried
state $Q_t \in \mathbb{R}^{m \times d}$ that threads through the stream via a
gated recurrence (Figure~\ref{fig:arch}b). At each step:
\emph{(1)}~the Perceiver encoder reads $S_t$ using the carried state $Q_{t-1}$ as
its latent queries, conditioning the read on all prior sessions---each slot pools
independently, attending only to the session tokens;
\emph{(2)}~a single self-attention layer over the $m$ slots lets them exchange
information and reorganise---this is the only point at which slots communicate;
\emph{(3)}~a gated write folds the result into the carried state:
\begin{align}
  g_t &= \sigma\bigl(W_g[x_t;\,Q_{t-1}] + b_g\bigr), \notag\\
  Q_t &= \mathrm{LN}\bigl((1-g_t)\,Q_{t-1} + g_t\,x_t\bigr).
  \label{eq:gate}
\end{align}
where $x_t$ is the output of the read-and-reorganise steps. The gate produces one
scalar per slot, so different slots can retain stable long-lived facts while others
are overwritten by new evidence. LayerNorm after each update prevents magnitude
drift over long streams.

\paragraph{Decode.}
A single Perceiver decoder block maps the final workspace $Q_N$ to adapter
weights. It uses $L \times r$ output queries---one per (layer, rank-index)
pair---each decoded into one rank-1 component:
\begin{equation}
  \Delta W_\ell \;=\; \alpha\, B_\ell A_\ell, \qquad
  A_\ell \in \mathbb{R}^{r \times d_{\mathrm{ff}}},\;\;
  B_\ell \in \mathbb{R}^{d \times r},
  \label{eq:lora}
\end{equation}
applied to the feed-forward down-projection in every block. The adapter rank is
fixed regardless of $N$: accumulation happens inside the carried state, not by
growing or merging adapters. Consolidating one additional session is an $O(1)$
update, and the persisted artifact is the same size.

\subsection{Episodic Buffer}

A fixed-size parametric state cannot retain every verbatim detail with perfect
fidelity---exact dates, proper names, precise figures require lossless surface form.
The episodic buffer provides this complement: it supplies local evidence to a model
whose weights have already been shaped by the full interaction history. The adapter
encodes cross-session structure---what the user cares about, how entities relate,
which topics recur---while the buffer provides the specific tokens the adapted model
needs to ground its answer.

We implement the buffer with minimal machinery: each session is embedded with a
sentence encoder \cite{reimers2019sbert}, and at query time we return the top-$k$ pieces by cosine
similarity. Our ablations show the gains are robust to retriever sophistication, and
stronger indexing is orthogonal and only helps
further.

\subsection{Inference}
\label{sec:inference}

At query time: \emph{(1)}~load the cached adapter
$\dw = H_\phi(Q_N)$;\; \emph{(2)}~retrieve top-$k$ spans $C$;\;
\emph{(3)}~merge $\theta' = \theta + \dw$;\;
\emph{(4)}~generate $y = f_{\theta'}(q, C)$.
Step~1 is amortised across all queries for a given user and requires no LLM calls
or gradient steps; step~2 is an $O(N)$ embedding comparison; step~3 is a free
weight merge; step~4 is standard generation.  The parametric channel is $O(1)$ in the number of past sessions: a constant-size state, a constant-cost update per session, and a constant-size adapter.

\subsection{Training}
\label{sec:training}

\method\ is trained by \emph{context distillation}: a teacher model answers each
query with the full session history in context, and the student---the frozen base
wearing only the generated adapter, without that history---must reproduce the
teacher's answer:
\begin{equation}
  \mathcal{L}(\phi) \;=\; \mathbb{E}_{(\Sset,q,y)\sim\mathcal{D}}
  \bigl[-\log\, p_{\theta + H_\phi(\Sset)}(y \mid q)\bigr],
  \label{eq:loss}
\end{equation}
a cross-entropy loss on teacher-generated answers with a small adapter-magnitude
penalty. Only the hypernetwork parameters $\phi$ are updated; the base model
remains frozen. Training uses no retrieval, so the adapter alone must carry the
answer---this forces the consolidation channel to be self-sufficient. Training proceeds in two stages. \textbf{Stage~1} initialises from a released
single-document-to-adapter checkpoint~\cite{charakorn2026doctolora}; the recurrent
machinery (gate, self-attention, positional embedding) is added as no-ops that
preserve the pre-trained behaviour. \textbf{Stage~2} teaches accumulation and
gating on multi-session sequences spanning up to several dozen sessions, using
truncated backpropagation through time to bound activation memory while still
training the recurrence over many steps.

\section{Training Data Generation}

\begin{figure}[tbp]
\centering
\includegraphics[width=\columnwidth]{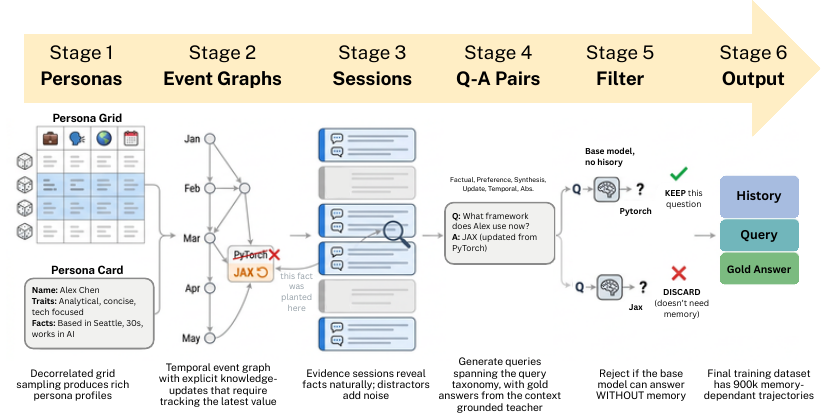}
\caption{The synthetic data pipeline: personas condition causal event
graphs with  knowledge-updates; filter questions unanswerable without history.}
\label{fig:pipe}
\end{figure}

No public dataset provides multi-session, memory-dependent training data at scale
(LoCoMo has $\sim$10 conversations, LongMemEval a few hundred questions), so we
build one with a fully automated six-stage pipeline (Figure~\ref{fig:pipe}), driven
throughout by Llama-3.3-70B-Instruct \cite{grattafiori2024llama3}. It follows the LongMemEval \cite{wu2025longmemeval} and LoCoMo
\cite{maharana2024locomo} recipes (planted evidence, distractor sessions, memory
taxonomies) but is automated end-to-end.

The pipeline runs in six stages (Figure~\ref{fig:pipe}). \textbf{(1)~Personas:}
thousands of personas are sampled from a balanced, decorrelated attribute grid
(industry, style, expertise, region, coherent age/life-stage, interests, plus a
``seed twist'' and cliche bans). \textbf{(2)~Event graphs:} each persona conditions
a temporal event graph of about two dozen causally linked events, several of them
\emph{knowledge-updates} that contradict an earlier fact, so a system must resolve
to the latest value. \textbf{(3)~Sessions:} the events are dramatized as multi-turn
conversations that reveal facts naturally, never as lists, across diverse session
types. \textbf{(4)~Distractors:} inter-persona distractor sessions are interleaved
to depths of over a hundred, so evidence is sparse within a long history.
\textbf{(5)~Questions:} at several history checkpoints we generate a balanced set of
six memory-type questions, factual, preference, cross-session synthesis,
knowledge-update, temporal, and abstention, with gold answers from a full-context
teacher. \textbf{(6)~Memory filter:} a \emph{must-require-memory} test discards any
question the base model can already answer with no history (keeping $\sim$82\%), so
the training signal is genuinely memory-dependent. An automated LLM-judge quality
pass runs between stages.
\begin{table}[tbp]
\centering
\resizebox{0.86\columnwidth}{!}{%
\small
\begin{tabular}{lr}
\toprule
Property & Value \\
\midrule
Trajectories generated & $\sim$900K \\
Distinct personas & 23{,}984 \\
History chunks per trajectory: median / range & 14 / 2--48 \\
Messages per chunk: median / range & 7 / 1--25 \\
History length (tokens): median / p90 & 5{,}457 / 13{,}322 \\
Memory types (balanced) & 6 \\
\midrule
\emph{vs.} LoCoMo (conversations) & 10 \\
\emph{vs.} LongMemEval-S (questions) & 500 \\
\bottomrule
\end{tabular}}
\caption{Synthetic corpus statistics and scale.}
\label{tab:data}
\end{table}


\begin{table*}[h]
\centering
\resizebox{\textwidth}{!}{%
\small
\setlength{\tabcolsep}{3.5pt}
\begin{tabular}{l ccccc >{\scriptsize}c | ccccccc >{\scriptsize}c}
\toprule
& \multicolumn{6}{c|}{\textbf{LoCoMo}} & \multicolumn{8}{c}{\textbf{LongMemEval-S}} \\
\cmidrule(lr){2-7}\cmidrule(lr){8-15}
\textbf{System} & Single & Multi & Temp. & Open & \textbf{Overall} & \normalsize $\Delta$\,[95\% CI] & SS$_u$ & SS$_a$ & Pref. & M-sess & K-upd. & Temp. & \textbf{Overall} & \normalsize $\Delta$\,[95\% CI] \\
\midrule
\multicolumn{15}{c}{\cellcolor{black!7}\textit{Base model: Gemma-2-2B}}\\
Mem0 & 50.6 & \tbest{45.4} & 19.6 & 29.2 & \second{41.8} & $+6.0\,[2.1,\,10.3]$ & 46.4 & 20.0 & 20.0 & 25.9 & 34.5 & 7.1 & 25.6 & $+14.0\,[4.7,\,23.3]$ \\
A-Mem & 32.2 & 34.4 & 15.6 & 26.0 & 28.7 & $+19.0\,[14.4,\,23.4]$ & 34.5 & 56.7 & \second{20.0} & 11.1 & 24.1 & \tbest{25.0} & 28.9 & $+10.4\,[1.7,\,19.1]$ \\
Generative Agents & 15.4 & 16.0 & 2.5 & 12.5 & 12.6 & $+35.1\,[31.4,\,38.3]$ & 37.9 & 30.0 & 10.0 & 0.0 & \tbest{65.5} & 3.6 & 24.9 & $+14.5\,[5.8,\,23.1]$ \\
MemoryBank & 11.9 & 12.1 & 3.7 & 14.6 & 10.4 & $+37.3\,[32.2,\,42.4]$ & 51.7 & \second{60.0} & \tbest{23.3} & 11.1 & 10.3 & 10.7 & 28.3 & $+11.0\,[2.3,\,19.7]$ \\
SimpleMem & 11.9 & 15.6 & 3.4 & 9.4 & 10.7 & $+37.1\,[33.1,\,41.2]$ & 0.0 & 10.0 & 3.3 & 0.0 & 0.0 & 0.0 & 2.3 & $+37.0\,[29.5,\,44.5]$ \\
\cmidrule(l){1-15}
\rowcolor{blue!4} \method\ \textbf{(ours)} & \tbest{58.6} & \second{38.7} & \tbest{27.4} & \tbest{49.0} & \tbest{47.7} & --- & \tbest{62.1} & 60.0 & 16.7 & \tbest{25.9} & 51.7 & \second{17.9} & \tbest{39.3} & --- \\
\;\; consolidation-only & 28.1 & 18.8 & 5.3 & \second{33.3} & 21.9 & $+25.8\,[22.9,\,29.1]$ & 13.8 & 16.7 & 16.7 & \second{14.8} & 17.2 & 14.3 & 15.6 & $+23.7\,[15.6,\,31.8]$ \\
\;\; episodic-only & \second{53.5} & 31.6 & \second{19.6} & 26.0 & 40.6 & $+7.1\,[5.7,\,9.1]$ & \second{55.2} & \tbest{63.3} & 13.3 & 3.7 & \second{62.1} & 10.7 & \second{35.3} & $+4.0\,[-2.9,\,11.6]$ \\
\midrule
\multicolumn{15}{c}{\cellcolor{black!7}\textit{Base model: Qwen3-4B}}\\
Mem0 & \second{61.3} & \tbest{58.2} & 22.4 & 26.0 & \second{50.4} & $+1.2\,[-1.5,\,4.0]$ & \tbest{75.9} & 53.3 & \second{50.0} & \tbest{40.7} & \second{62.1} & \second{14.3} & \tbest{49.7} & $-6.4\,[-16.2,\,3.5]$ \\
A-Mem & 48.0 & 40.4 & 21.2 & 13.5 & 38.8 & $+12.8\,[8.2,\,16.9]$ & 13.8 & 26.7 & 26.7 & 0.0 & 6.9 & 0.0 & 12.7 & $+30.6\,[22.0,\,39.3]$ \\
Generative Agents & 14.9 & 11.7 & 4.4 & 6.2 & 11.6 & $+40.0\,[36.4,\,43.4]$ & 24.1 & 36.7 & 6.7 & 3.7 & 44.8 & 3.6 & 20.2 & $+23.1\,[14.5,\,31.2]$ \\
MemoryBank & 9.5 & 10.3 & 2.5 & 8.3 & 8.1 & $+43.5\,[40.1,\,46.9]$ & 48.3 & 53.3 & 33.3 & 14.8 & 13.8 & 7.1 & 28.9 & $+14.5\,[5.8,\,23.1]$ \\
SimpleMem & 1.8 & 25.2 & 13.4 & 19.8 & 9.7 & $+41.9\,[36.8,\,47.1]$ & 34.5 & 63.3 & \tbest{60.0} & 7.4 & 24.1 & 10.7 & 34.1 & $+9.2\,[0.6,\,17.9]$ \\
\cmidrule(l){1-15}
\rowcolor{blue!4} \method\ \textbf{(ours)} & \tbest{61.8} & \second{45.7} & \tbest{31.8} & \tbest{46.9} & \tbest{51.6} & --- & 48.3 & \tbest{80.0} & 33.3 & \second{25.9} & 55.2 & 14.3 & \second{43.4} & --- \\
\;\; consolidation-only & 27.7 & 18.8 & 5.9 & \second{35.4} & 22.0 & $+29.6\,[26.4,\,33.0]$ & 13.8 & 13.3 & 20.0 & 22.2 & 10.3 & 10.7 & 15.0 & $+28.3\,[20.2,\,36.4]$ \\
\;\; episodic-only & 51.6 & 31.6 & \second{22.7} & 32.3 & 40.6 & $+11.0\,[9.5,\,12.6]$ & \second{51.7} & \second{66.7} & 40.0 & 11.1 & \tbest{62.1} & \tbest{14.3} & 41.6 & $+1.7\,[-5.2,\,8.7]$ \\
\midrule
\multicolumn{15}{c}{\cellcolor{black!7}\textit{Base model: Mistral-7B}}\\
Mem0 & 53.0 & \second{46.5} & 17.4 & 35.4 & 43.2 & $+12.0\,[8.1,\,16.2]$ & 41.4 & 46.7 & \tbest{56.7} & 7.4 & 27.6 & 14.3 & 32.9 & $+20.2\,[11.0,\,29.5]$ \\
A-Mem & 51.3 & 40.8 & 18.1 & 27.1 & 40.9 & $+14.4\,[8.0,\,20.1]$ & 10.3 & 33.3 & 36.7 & 0.0 & 6.9 & 0.0 & 15.0 & $+38.2\,[29.5,\,46.8]$ \\
Generative Agents & 13.5 & 15.6 & 2.8 & 14.6 & 11.7 & $+43.6\,[38.7,\,48.1]$ & 24.1 & 46.7 & 30.0 & 0.0 & \second{65.5} & 7.1 & 29.5 & $+23.7\,[15.0,\,31.8]$ \\
MemoryBank & 11.6 & 11.7 & 2.8 & 13.5 & 9.9 & $+45.4\,[40.1,\,51.0]$ & 48.3 & 60.0 & \second{53.3} & 7.4 & 10.3 & 7.1 & 31.8 & $+21.4\,[12.7,\,30.1]$ \\
SimpleMem & 12.8 & 19.9 & 6.9 & 18.8 & 13.2 & $+42.1\,[29.5,\,53.2]$ & 10.3 & 20.0 & 30.0 & 0.0 & 0.0 & 3.6 & 11.0 & $+41.9\,[33.1,\,50.0]$ \\
\cmidrule(l){1-15}
\rowcolor{blue!4} \method\ \textbf{(ours)} & \tbest{64.0} & \tbest{47.5} & \tbest{39.6} & \tbest{55.2} & \tbest{55.3} & --- & \tbest{69.0} & \tbest{80.0} & 40.0 & \tbest{33.3} & \tbest{69.0} & \tbest{25.0} & \tbest{53.2} & --- \\
\;\; consolidation-only & 28.8 & 19.9 & 10.3 & \second{37.5} & 23.8 & $+31.5\,[26.1,\,36.3]$ & 17.2 & 13.3 & 23.3 & 7.4 & 17.2 & 14.3 & 15.6 & $+37.6\,[28.9,\,46.2]$ \\
\;\; episodic-only & \second{59.0} & 44.7 & \second{26.8} & 29.2 & \second{47.7} & $+7.5\,[5.3,\,9.9]$ & \second{62.1} & \second{66.7} & 50.0 & \second{11.1} & 62.1 & \second{25.0} & \second{46.8} & $+6.4\,[0.0,\,12.7]$ \\
\bottomrule
\end{tabular}}
\caption{\textbf{Agentic (token-space) memory}, LLM-judge accuracy (\%) by sub-category.
$\Delta$ is \method\ minus that row on the paired items, with a 95\% bootstrap interval.}
\label{tab:main}
\end{table*}

\begin{table*}[h]
\centering
\resizebox{\textwidth}{!}{%
\small
\setlength{\tabcolsep}{4pt}
\renewcommand{\arraystretch}{1.1}
\begin{tabular}{l c ccc >{\scriptsize}c >{\scriptsize}c ccc >{\scriptsize}c >{\scriptsize}c}
\toprule
& & \multicolumn{5}{c}{\textbf{LoCoMo}} & \multicolumn{5}{c}{\textbf{LongMemEval-S}} \\
\cmidrule(lr){3-7}\cmidrule(lr){8-12}
\textbf{System} & \textbf{rank} & \textsc{G} & \textsc{Q} & \textsc{M} & \normalsize $\Delta$ vs \method & \normalsize $\Delta$ vs consol.
 & \textsc{G} & \textsc{Q} & \textsc{M} & \normalsize $\Delta$ vs \method & \normalsize $\Delta$ vs consol. \\
\midrule
D2L-Merge & 8 per session & 1.1 & 4.4 & 1.1 & $+44.6\,[40.5,\,49.0]$ & $+18.0\,[14.9,\,21.2]$ & 0.6 & 1.2 & 1.2 & $+44.3\,[39.9,\,48.7]$ & $+14.5\,[11.4,\,17.7]$ \\
D2L-Retrieve & 8 & 16.1 & 15.0 & 18.9 & $+30.2\,[24.1,\,36.1]$ & $+3.5\,[-0.8,\,7.9]$ & 12.1 & 11.0 & 12.7 & $+33.3\,[28.5,\,38.2]$ & $+3.5\,[-0.2,\,7.1]$ \\
Test-time LoRA\ttt & 8 & 1.1 & 21.7 & 8.9 & $+36.3\,[31.7,\,40.9]$ & $+9.6\,[5.5,\,13.8]$ & 1.7 & 6.4 & 2.9 & $+41.6\,[37.2,\,46.1]$ & $+11.8\,[8.5,\,15.0]$ \\
TLM\ttt & 8 & 1.1 & 3.3 & 3.3 & $+44.3\,[40.1,\,48.5]$ & $+17.6\,[14.4,\,21.0]$ & 1.2 & 7.5 & 12.1 & $+38.3\,[33.7,\,43.0]$ & $+8.5\,[5.0,\,11.9]$ \\
TMEM-once\ttt & 8 & 4.4 & 10.0 & 5.6 & $+40.2\,[35.9,\,45.0]$ & $+13.5\,[10.1,\,17.4]$ & 4.6 & 9.2 & 8.1 & $+38.0\,[33.3,\,42.6]$ & $+8.1\,[4.6,\,11.8]$ \\
TMEM-per-question\ttt & 8 & 15.6 & \tbest{43.3} & 17.8 & $+21.3\,[16.2,\,26.5]$ & $-5.4\,[-9.3,\,-0.9]$ & 3.5 & \second{36.4} & 9.8 & $+28.7\,[24.1,\,33.3]$ & $-1.2\,[-5.2,\,2.9]$ \\
FT-Ceiling\ttt & full & \second{33.3} & 38.9 & \second{36.7} & $+10.6\,[5.1,\,16.2]$ & $-16.1\,[-19.6,\,-12.3]$ & 5.2 & 7.5 & 6.4 & $+38.9\,[34.3,\,43.5]$ & $+9.1\,[5.8,\,12.3]$ \\
\method\ consolidation-only & 8 & 20.6 & 21.7 & 18.3 & $+26.7$ & --- & \second{15.6} & 15.0 & \second{15.6} & $+29.9\,[25.0,\,34.5]$ & --- \\
\rowcolor{blue!7} \method\ \textbf{(ours)} & 8 & \tbest{48.3} & \second{41.7} & \tbest{50.6} & --- & & \tbest{39.3} & \tbest{43.4} & \tbest{53.2} & --- & \\
\bottomrule
\end{tabular}}
\caption{\textbf{Weight-space (parametric) memory}, judge accuracy (\%) from the adapter
alone. \ttt\ = \textbf{trains at test time} (takes gradient steps on the conversation before
it can answer); unmarked rows are frozen. $\Delta$ is \method's pooled advantage over that
row, with a 95\% bootstrap interval.}
\label{tab:parametric}
\end{table*}

\begin{table*}[h]
\centering
\resizebox{0.8\textwidth}{!}{%
\small
\setlength{\tabcolsep}{5pt}
\renewcommand{\arraystretch}{1.1}
  \begin{tabular}{l rrrr rrr}
  \toprule
  & \multicolumn{4}{c}{\textbf{Building the memory} \scriptsize(once, per conversation)} & \multicolumn{3}{c}{\textbf{Answering} \scriptsize(per question)}\\
  \cmidrule(lr){2-5}\cmidrule(lr){6-8}
  \textbf{System} & LLM decode calls & LLM tokens & grad.\ steps & wall (s) & LLM tokens & LLM decode calls & wall (s)\\
  \midrule
  Mem0              & 48.8  & $4.9{\times}10^{5}$ & 0 & 200.6     & 778      & 1.00 & 5.84\\
  A-Mem             & 464.1 & $6.2{\times}10^{5}$ & 0 & 1{,}554.5 & 1{,}396  & 1.00 & 5.07\\
  Generative Agents & 34.7  & $5.1{\times}10^{4}$ & 0 & 34.8      & 688      & 1.00 & 4.94\\
  MemoryBank        & 28.2  & $3.0{\times}10^{4}$ & 0 & 59.7      & 928      & 1.00 & 5.31\\
  SimpleMem         & 12.1  & $1.0{\times}10^{5}$ & 0 & 109.2     & 16{,}016 & 5.62 & 12.30\\
  \midrule
  Test-time LoRA\ttt      & 0   & 0                   & 17.7  & 14.5  & 50  & 1.00 & 2.64\\
  TLM\ttt                 & 0   & 0                   & 62.6  & 25.2  & 65  & 1.00 & 4.04\\
  TMEM-once\ttt           & 5.9 & $2.9{\times}10^{4}$ & 31.0  & 175.0 & 114 & 1.00 & 4.03\\
  FT-Ceiling\ttt          & 0   & 0                   & 47.2  & 49.5  & 50  & 1.00 & 1.86\\
  \rowcolor{red!7} TMEM-per-question\ttt & 107.5 & $5.2{\times}10^{5}$ & 315.0 & 459.0 & 73 & 1.00 & 4.03\\
  \midrule
  \rowcolor{blue!4} \method\ \textbf{(ours)}, consolidation only & 0 & 0 & 0 & 10.3 & 70      & 1.00 & 3.25\\
  \rowcolor{blue!4} \method\ \textbf{(ours)}                     & 0 & 0 & 0 & 10.3 & 1{,}583 & 1.00 & 3.25\\
  \bottomrule
  \end{tabular}}
\caption{\textbf{Cost of each memory}, on Qwen3-4B over LoCoMo: build per conversation, answer per
question. \ttt\ = gradient steps at test time; the red row rebuilds its memory per question, and its
build columns give the cost of one rebuild.}
\label{tab:efficiency}
\end{table*}

\section{Experimental Setup}

\subsection{Benchmarks}
We evaluate on \textbf{LoCoMo} \cite{maharana2024locomo} and \textbf{LongMemEval-S} \cite{wu2025longmemeval}, which are standard LLM memory evaluation benchmarks. On LoCoMo we use the 1,540 questions of the four answerable categories (the adversarial category is excluded). For LongMemEval-S we sample 30 questions per type (seed 0) and evaluate on the 173 answerable ones; its 7 abstention items are excluded, as LoCoMo's adversarial category is. Because the parametric baselines train at test time, they are compared on a 180-question LoCoMo subset and on the same LongMemEval-S questions.
Every system is run on the same three open base models: \textbf{Gemma-2-2B} \cite{gemmateam2024gemma2}, \textbf{Qwen3-4B}
\cite{qwen2025qwen3} and \textbf{Mistral-7B} \cite{jiang2023mistral}. On both benchmarks the episodic buffer retrieves the same unit: a session split on turn boundaries into pieces of at most 1,800 characters.

\subsection{Baselines}
\label{sec:baselines}

\paragraph{Agentic (token-space).} Systems that store the history as text and retrieve it at
question time: \textbf{Mem0} \cite{chhikara2025mem0}, \textbf{A-Mem} \cite{xu2025amem},
\textbf{Generative Agents} \cite{park2023generative}, \textbf{MemoryBank}
\cite{zhong2024memorybank} and \textbf{SimpleMem} \cite{simplemem2026}, each run with its
published pipeline.

\paragraph{Parametric (weight-space).} Systems that write the history into parameters, our own
setting. We reimplement \textbf{TMEM} \cite{tmem2026} in both of its modes, built once per
conversation (\textbf{TMEM-once}) (where the model is prompted to generate training data based on conversational history, agnostic of the current query) and rebuilt per question (\textbf{TMEM-per-question}), and
\textbf{TLM} \cite{hu2025tlm} (where the model is trained for each new query). We construct four strong baselines: \textbf{Test-time LoRA}, the same rank-8
adapter on the same projection fitted by SGD; \textbf{FT-Ceiling}, a full fine-tune of every
weight under one protocol for all bases (AdamW \cite{loshchilov2019adamw}, batch 4, 3\% warmup, clipping 1.0, lr
$5\times10^{-5}$ and 8 epochs on LoCoMo, 2 on LongMemEval-S;); and \textbf{D2L-Merge} and
\textbf{D2L-Retrieve}, the same released hypernetwork \cite{charakorn2026doctolora}, with the first based on their proposed chunk concatenation mechanism, and the second involving retrieving first to reduce the tokens to be internalized.


\section{Results}


\subsection{Memory at Zero Construction Cost Is Competitive}
\label{sec:main}

\noindent\textbf{Claim.} \method\ leads every agentic memory system on LoCoMo, and on LongMemEval-S on two of three bases, trailing only Mem0 on Qwen by a non-significant margin, while building its memory with no generative language-model calls, no gradient steps, and no generated tokens.

\begin{figure}[t]
\centering
\includegraphics[width=\columnwidth]{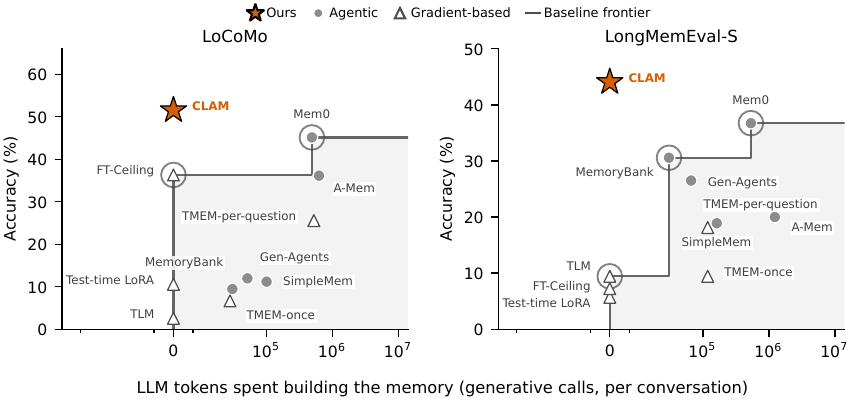}
\caption{Pareto frontier of cost versus quality on both benchmarks (LoCoMo and LongMemEval-S).}
\label{fig:pareto}
\end{figure}

Table~\ref{tab:main} reports accuracy and Table~\ref{tab:efficiency} the corresponding costs. Each row reports \method\ minus that system on paired items, with a 95\% bootstrap interval. Figure~\ref{fig:pareto} plots the cost--accuracy trade-off, on which \method\ is the most accurate system at zero construction cost. On LoCoMo \method\ leads the strongest agentic system (Mem0) by 6.0 on Gemma, 1.2 on Qwen, and 12.0 on Mistral; no other agentic system comes within 12 points on any base. On LongMemEval-S \method\ leads every agentic baseline on Gemma (by 10.4 to 37.0) and Mistral (by 20.2 to 41.9) and only trails Mem0 on Qwen ($-6.4$, not significant). That gap reflects a favourable interaction between Qwen's strong agentic abilities and Mem0's extraction pipeline, but it comes at 48.8 generative LLM calls and $4.9\times 10^{5}$ LLM tokens per conversation, a budget \method\ never spends. The sub-category columns show that \method's LoCoMo gains concentrate in \emph{temporal} and \emph{open-domain} questions---the categories whose evidence no retriever surfaces verbatim--- showing efficacy of the parametric consolidation stream.

\begin{table}[t]
\centering
\resizebox{\columnwidth}{!}{%
\small
\setlength{\tabcolsep}{5pt}
\renewcommand{\arraystretch}{1.1}
\begin{tabular}{l ccc}
\toprule
\textbf{Configuration} & \textsc{G} & \textsc{Q} & \textsc{M} \\
\midrule
\multicolumn{4}{l}{\cellcolor{black!5}\emph{(a) Single stream vs.\ both streams}}\\
Consolidation only \scriptsize(no retrieval) & 21.9 & 22.0 & 23.8 \\
Episodic only \scriptsize(no adapter, \emph{ref.}) & 40.6 & 40.6 & 47.7 \\
\rowcolor{blue!4} \textbf{Both (\method)}    & \tbest{47.7} & \tbest{51.6} & \tbest{55.3} \\
\midrule
\multicolumn{4}{l}{\cellcolor{black!5}\emph{(b) Consolidation gate}}\\
Reset state \scriptsize(carried state discarded every session) & 16.7 & 17.2 & \tbest{24.1} \\
Ungated write \scriptsize($g_t{\equiv}1$, overwrite)           & 21.4 & 21.2 & 22.5 \\
\rowcolor{blue!4} \textbf{Learned gate (\method)}           & \tbest{21.9} & \tbest{22.0} & 23.8 \\
\midrule
\multicolumn{4}{l}{\cellcolor{black!5}\emph{(c) Is it the adapter, or \emph{this} adapter?}}\\
No adapter \scriptsize(base priors, no memory of any kind) & 1.1 & 3.9 & 2.8 \\
Untrained adapter \scriptsize(same shape, random weights)  & 0.0 & 0.6 & 0.0 \\
\rowcolor{blue!4} \textbf{\method's adapter}                    & \tbest{20.6} & \tbest{21.7} & \tbest{18.3} \\

\bottomrule
\end{tabular}}
\caption{\textbf{Ablations} (LoCoMo, judge accuracy \%): (a) one stream versus both, (b) the
consolidation gate, (c) whether any adapter would do.}
\label{tab:ablation}
\end{table}

\begin{figure}[t]
\centering
\includegraphics[width=0.8\columnwidth]{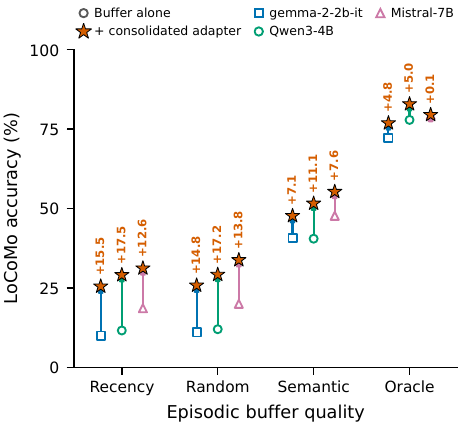}
\caption{\textbf{The adapter helps at every buffer quality} (LoCoMo). Open markers are the
buffer alone; stars add the consolidated adapter.}
\label{fig:buffer}
\end{figure}

\begin{figure}[t]
\centering
\includegraphics[width=0.8\columnwidth]{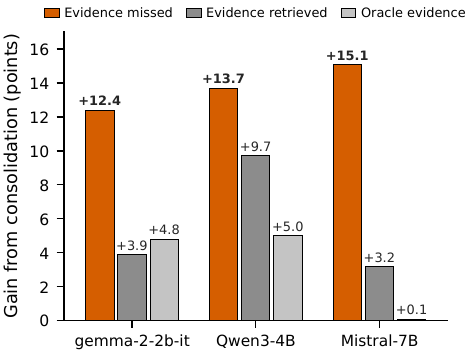}
\caption{\textbf{The adapter pays when retrieval fails} (LoCoMo). Gain from adding the
adapter, split by whether the buffer retrieved the gold evidence.}
\label{fig:hitmiss}
\end{figure}


\subsection{Both Channels Are Necessary and Complementary}
\label{sec:ablation}

\noindent\textbf{Claim.} Neither channel alone reproduces the full system, and the two are complementary: the adapter carries information retrieval \emph{cannot} supply.

\paragraph{Single-stream ablation.} Block~(a) of Table~\ref{tab:ablation} runs each stream on its own. Adding the adapter on top of retrieval adds 7.1, 11.0, and 7.5 points on Gemma, Qwen, and Mistral; adding retrieval on top of the adapter adds 25.8 to 31.5 (all intervals conversation-clustered and clear of zero). 

\paragraph{Oracle buffer.} Figure~\ref{fig:buffer} varies the buffer from degenerate to ideal. With an oracle buffer, which places the gold evidence in context, redundancy predicts no gain, yet the adapter adds $+4.8$ [2.2, 7.1] on Gemma and $+5.0$ [2.8, 7.2] on Qwen (both $p{<}10^{-3}$); on Mistral, already at 79\%, it ties ($+0.1$). The weights carry information retrieval cannot supply, cross-session integration, implicit patterns, accumulated preferences, rather than a better-indexed copy of it.

 \paragraph{When retrieval fails.} Figure~\ref{fig:hitmiss} shows the converse: on the 36\% of questions whose evidence the buffer misses, the adapter adds $+12.4$, $+13.7$ and $+15.1$ points, against $+3.9$, $+9.7$ and $+3.2$ where retrieval succeeds. The adapter's contribution is largest exactly where the episodic channel breaks down, and it remains positive when the evidence is already in context, including under the oracle buffer: the consolidated weights help the model use the evidence it is given, not only recover what retrieval missed.

\paragraph{Controls.} Discarding the carried state at every session costs 4.8--5.2 points on Gemma and Qwen (Table~\ref{tab:ablation}b), and a random adapter of the same shape scores 0.0--0.6\% (c). 


\subsection{Parametric Consolidation Scales Where Test-Time Training Does Not}
\label{sec:parametric}

\noindent\textbf{Claim.} The full \method\ system leads every weight-space memory; its consolidation channel alone leads every rank-8 method that adapts once per conversation, without a gradient step, and holds its accuracy on long histories.

\begin{figure}[t]
\centering
\includegraphics[width=0.8\columnwidth]{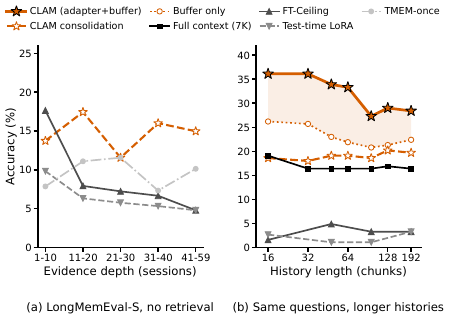}
\caption{\textbf{Accuracy as the history grows.} (a) By distance between evidence and question (LongMemEval-S, no retrieval). (b) On the same set of questions, with growing history length.}
\label{fig:history}
\end{figure}

Table~\ref{tab:parametric} evaluates each baseline as designed, without retrieval. The full system leads every method by 10.6 to 44.6 points on LoCoMo and 28.7 to 44.3 on LongMemEval-S. Like for like, the consolidation channel leads test-time LoRA, TLM, TMEM-once and both D2L variants on both benchmarks; only FT-Ceiling and TMEM-per-question, which rebuilds its memory for every question (315 gradient steps and 107.5 LLM calls per rebuild), lead it, and only on LoCoMo. Given the same episodic buffer, \method\ leads all five weight-space baselines on LoCoMo ($+6.9$ to $+15.7$).

On LongMemEval-S's 4.5$\times$ longer histories, FT-Ceiling falls to 5.2--7.5 while the consolidation channel holds 15.0--15.6 ($+9.1$ [5.8, 12.3]). Figure~\ref{fig:history}(a) locates the failure: FT-Ceiling loses 2.2 points per 10 sessions of evidence depth, while \method\ stays flat. In a controlled sweep on our synthetic validation set from 16 to 192 chunks (Figure~\ref{fig:history}(b)), the consolidation channel stays level, the adapter adds 6 to 11 points over the buffer at every length, and test-time training stays below 5\%.

\section{Conclusion}
Existing LLM memory systems consolidate experience only in token space. We introduced
\method, which adds the missing half: a recurrent hypernetwork compiles a stream of
sessions into a low-rank adapter in a single forward pass, with no gradient steps and no
generation calls, while a gated latent state carries knowledge forward in fixed size and a
lightweight retrieval channel supplies verbatim detail at inference. Across two benchmarks
and three open base models the two channels prove complementary, and the memory is built
at a cost no comparable system approaches. Weight-space consolidation is a practical second
memory channel for language agents, and we hope it encourages work on memory that is
internalised rather than re-read.

\section*{Limitations}
\method\ is evaluated on two English conversational-memory benchmarks with base models of
2--7\,B parameters; behavior at larger scale, in other languages, and on non-conversational
histories such as documents or code remains to be established. The consolidated adapter is a
fixed-size store, though better results can be expected from scaling the adapter rank and the
hypernetwork's parameter count. Finally, the hypernetwork is trained
once per base model on synthetic multi-session data; transferring a trained hypernetwork to an
unseen base model without retraining is outside the scope of this work.

\section*{Ethics Statement}
This work uses two public benchmarks, LoCoMo and LongMemEval-S, together with a synthetic
corpus generated from fictional personas. Both benchmarks, the open base models and the
baseline implementations are used in accordance with their respective licences and intended
research use; no human subjects were involved and no personal data was collected. 
\bibliography{custom}

\appendix

\section{Further Details}

\subsection{Implementation Details}
The context encoder is a Perceiver-style resampler of $9$ cross-attention blocks with $8$ latent
slots per layer, a $4\times$ feed-forward expansion, SiLU activations and RMSNorm. Keys and values
are the session activations alone, so slots pool a session independently and exchange no information
during a read; the only slot-to-slot mixing is the between-session consolidation block, an $8$-head
self-attention with pre- and post-LayerNorm and a $4\times$ expansion, applied once per session
boundary. A single-block decoder carries one output query per (layer, rank index) pair --- $L\times r$
queries, $208$ for a $26$-layer base at $r{=}8$ --- and each decodes into one rank-1 component. The
resulting adapter is rank-8 on \texttt{down\_proj} in every block, $\alpha{=}r$, dropout $0$:
$2.40$\,M / $3.54$\,M / $4.72$\,M parameters on Gemma / Qwen / Mistral, about $7$\,MB in half
precision, plus a fixed learned bias that does not vary with the conversation. The head is
zero-initialised, so an untrained hypernetwork is an exact no-op on the base model. The episodic
buffer uses all-MiniLM-L6-v2 over session pieces of at most 1,800 characters, cosine top-$k$ with $k{=}5$ and a $7$K-token
budget, re-sorted chronologically. At inference the adapter is merged into the base weights and the
answer is a single greedy pass capped at $48$ new tokens.

\subsection{Training}
Only the hypernetwork is trained, by context distillation against a teacher that answers with the
full history in context. Stage~1 initialises from the released document-to-adapter checkpoint with
the recurrence inserted as a no-op. Stage~2 trains the carried state on multi-session examples of
$3$--$30$ sessions with truncated backpropagation through time: cross-entropy on the gold answer
plus a small penalty on adapter magnitude, AdamW at $1\times10^{-5}$, gradient accumulation $8$ for
an effective batch of $64$, packed context and input lengths of $6144$ tokens, and a checkpoint every
$1{,}000$ steps. Reported checkpoints are $25$K Stage-2 steps for Gemma and $8$K for Qwen and
Mistral. Training runs on a single A100-80GB and is paid once per base
model, never per conversation.

\subsection{Evaluation}
\paragraph{Judge.} All reported accuracies use one grader, \texttt{llama3.3-70b-instruct} at
temperature $0$, with an identical prompt for every system and both benchmarks. The system prompt is:
\begin{quote}\footnotesize\ttfamily
You are a strict grader for a question-answering benchmark over long conversations. You are given a
QUESTION, the reference GOLD answer(s), and a model PREDICTION. Decide whether the PREDICTION is
factually correct: it must convey the same key fact(s) as the gold answer. Ignore differences in
phrasing, extra correct detail, or verbosity. If the prediction states it cannot answer / no
information, and the gold is a real answer, it is incorrect. Respond with EXACTLY one word: Yes or No.
\end{quote}
and the user prompt is \texttt{QUESTION: \{q\}} / \texttt{GOLD ANSWER(S): \{g\_1\} | \{g\_2\} | ...} /
\texttt{PREDICTION: \{p\}} / \texttt{Is the prediction factually correct and consistent with the GOLD ANSWER(S)? Answer Yes or No.}

\begin{figure}[t]
\centering
\includegraphics[width=\columnwidth]{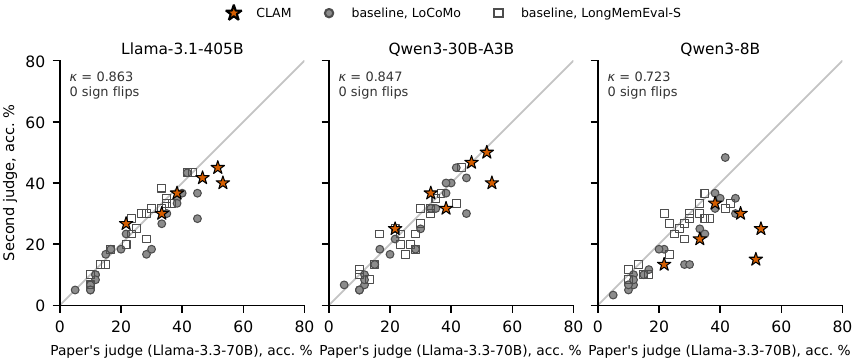}
\caption{Per-system accuracy under the paper's judge vs.\ each second judge.}
\label{fig:judges}
\end{figure}

\paragraph{Judge independence.} Because the judge shares a model family with the data generator, we
re-grade the same outputs with three other judges under the identical prompt: 2{,}880 judgments covering
360 items, 8 systems, both benchmarks and all three bases (Table~\ref{tab:judges},
Figure~\ref{fig:judges}). Agreement is high, and no \method-versus-baseline comparison changes sign and
significance under any judge.

\begin{table}[t]
\centering
\resizebox{\columnwidth}{!}{%
\small
\begin{tabular}{l c c c c}
\toprule
\textbf{Second judge} & \textbf{Agree} & \textbf{$\kappa$} & \textbf{Sign kept} & \textbf{Sig.\ flips}\\
\midrule
Llama-3.1-405B (same family, 5.8$\times$ larger) & 94.7\% & 0.863 & 39/42 & 0\\
Qwen3-30B-A3B (independent family)               & 94.1\% & 0.847 & 34/42 & 0\\
Qwen3-8B (independent family)                    & 89.7\% & 0.723 & 32/42 & 0\\
\bottomrule
\end{tabular}}
\caption{\textbf{Agreement with independent judges} on the same 2{,}880 judgments. \emph{Sig.\ flips}: \method-vs-baseline comparisons whose sign changes with significance.}
\label{tab:judges}
\end{table}

\paragraph{Metric.} Alongside judge accuracy we report \textbf{Cover-EM}
\cite{press2023compositionality,rosset2020knowledge}. Let $\mathrm{nrm}(\cdot)$ be the standard
answer normalisation (lowercasing, stripping punctuation and articles, collapsing whitespace), $p$ a
prediction and $\mathcal{G}$ its set of gold answers. Cover-EM scores a prediction $1$ if any gold
string occurs in it as a substring:
\begin{equation}
\mathrm{CoverEM}(p,\mathcal{G}) \;=\; \max_{g\in\mathcal{G}}\;
\mathbf{1}\!\left[\,\mathrm{nrm}(g)\subseteq\mathrm{nrm}(p)\,\right],
\label{eq:coverem}
\end{equation}
and the reported figure is its mean over questions. Because the criterion is containment, tokens the
prediction adds beyond the gold neither help nor hurt: it is invariant to answer length, unlike
token-F1, whose precision term falls as a prediction lengthens. This matters here because mean answer
length varies by an order of magnitude across systems.

\section{Further Results}

\subsection{Text-Grounded Evaluation}
Tables~\ref{tab:mainCEM} and~\ref{tab:paramCEM} repeat the two main results tables under Cover-EM,
with identical layout and identical statistical treatment. The ordering is broadly preserved, which
indicates the headline result does not depend on the LLM grader.

\begin{table*}[t]\centering
\resizebox{\textwidth}{!}{%
\small\setlength{\tabcolsep}{3.5pt}\renewcommand{\arraystretch}{1.1}
\begin{tabular}{l ccccc >{\scriptsize}c | ccccccc >{\scriptsize}c}
\toprule
& \multicolumn{6}{c|}{\textbf{LoCoMo}} & \multicolumn{8}{c}{\textbf{LongMemEval-S}} \\
\cmidrule(lr){2-7}\cmidrule(lr){8-15}
\textbf{System} & Single & Multi & Temp. & Open & \textbf{Overall} & \normalsize $\Delta$\,[95\% CI] & SS$_u$ & SS$_a$ & Pref. & M-sess & K-upd. & Temp. & \textbf{Overall} & \normalsize $\Delta$\,[95\% CI] \\
\midrule
\multicolumn{15}{c}{\cellcolor{black!7}\textit{Base model: Gemma-2-2B}}\\
Mem0 & 24.3 & 11.0 & 2.8 & 8.3 & 16.4 & $+5.2\,[3.4,\,7.3]$ & 39.3 & 13.3 & 0.0 & 18.5 & 27.6 & 7.1 & 17.4 & $+10.9\,[3.7,\,18.2]$ \\
A-Mem & 14.9 & 7.1 & \second{5.9} & 6.2 & 11.1 & $+10.5\,[7.9,\,13.0]$ & 34.5 & 26.7 & 0.0 & 7.4 & 13.8 & 10.7 & 15.6 & $+12.7\,[6.1,\,19.4]$ \\
Generative Agents & 5.3 & 3.5 & 1.9 & 3.1 & 4.1 & $+17.5\,[15.5,\,19.1]$ & 34.5 & 16.7 & 0.0 & 0.0 & \tbest{51.7} & 3.6 & 17.9 & $+10.4\,[3.2,\,17.1]$ \\
MemoryBank & 4.1 & 3.5 & 0.9 & 4.2 & 3.3 & $+18.2\,[16.1,\,20.4]$ & 37.9 & 26.7 & 0.0 & 3.7 & 10.3 & 7.1 & 14.5 & $+13.9\,[8.4,\,20.0]$ \\
SimpleMem & 3.7 & 3.5 & 1.2 & 2.1 & 3.1 & $+18.5\,[16.3,\,20.3]$ & 0.0 & 0.0 & 0.0 & 0.0 & 0.0 & 3.6 & 0.6 & $+27.7\,[21.6,\,34.3]$ \\
\cmidrule(l){1-15}
\rowcolor{blue!4} \method\ \textbf{(ours)} & \tbest{28.8} & \tbest{15.2} & \tbest{8.1} & \tbest{22.9} & \tbest{21.6} & --- & \tbest{55.2} & \second{36.7} & 0.0 & \second{11.1} & 41.4 & \tbest{25.0} & \tbest{28.3} & --- \\
\;\; consolidation-only & 4.6 & 8.5 & 2.8 & \second{15.6} & 5.6 & $+16.0\,[14.2,\,17.7]$ & 17.2 & 6.7 & \second{0.0} & \tbest{14.8} & 0.0 & \second{17.9} & 9.2 & $+19.1\,[12.5,\,25.8]$ \\
\;\; episodic-only & \second{28.4} & \second{11.3} & 2.5 & 9.4 & \second{18.6} & $+2.9\,[1.8,\,4.1]$ & \second{48.3} & \tbest{50.0} & \tbest{0.0} & 0.0 & \second{44.8} & 10.7 & \second{26.0} & $+2.3\,[-2.7,\,7.3]$ \\
\midrule
\multicolumn{15}{c}{\cellcolor{black!7}\textit{Base model: Qwen3-4B}}\\
Mem0 & 26.4 & \tbest{13.5} & 5.0 & 13.5 & 18.7 & $+2.9\,[0.3,\,5.6]$ & \tbest{62.1} & 40.0 & 0.0 & \tbest{25.9} & \second{51.7} & 17.9 & \tbest{32.9} & $-2.3\,[-10.1,\,5.5]$ \\
A-Mem & 26.3 & 10.6 & 4.7 & 5.2 & 17.5 & $+4.1\,[1.1,\,7.4]$ & 13.8 & 13.3 & 0.0 & 0.0 & 3.4 & 17.9 & 8.1 & $+22.5\,[15.8,\,29.1]$ \\
Generative Agents & 6.3 & 2.8 & 2.2 & 3.1 & 4.6 & $+17.0\,[14.9,\,19.2]$ & 20.7 & 23.3 & 0.0 & 0.0 & 37.9 & 0.0 & 13.9 & $+16.8\,[10.1,\,23.5]$ \\
MemoryBank & 3.5 & 3.5 & 0.6 & 2.1 & 2.8 & $+18.8\,[16.5,\,21.2]$ & 41.4 & 30.0 & 0.0 & 11.1 & 13.8 & 7.1 & 17.3 & $+13.3\,[6.6,\,19.9]$ \\
SimpleMem & 0.8 & 6.4 & \second{8.1} & 6.2 & 3.7 & $+17.9\,[15.5,\,20.2]$ & 27.6 & 33.3 & 0.0 & 7.4 & 17.2 & 7.1 & 15.6 & $+15.0\,[8.9,\,21.7]$ \\
\cmidrule(l){1-15}
\rowcolor{blue!4} \method\ \textbf{(ours)} & \tbest{29.2} & \second{13.1} & \tbest{9.3} & \tbest{21.9} & \tbest{21.6} & --- & 41.4 & \tbest{50.0} & 0.0 & 18.5 & 48.3 & \tbest{25.0} & \second{30.6} & --- \\
\;\; consolidation-only & 3.3 & 6.0 & 2.5 & \second{16.7} & 4.4 & $+17.1\,[15.0,\,19.3]$ & 3.4 & 0.0 & \second{0.0} & \second{22.2} & 3.4 & 17.9 & 7.5 & $+23.1\,[16.5,\,30.3]$ \\
\;\; episodic-only & \second{28.1} & 12.8 & 5.6 & 13.5 & \second{19.6} & $+2.0\,[0.7,\,3.1]$ & \second{44.8} & \second{40.0} & \tbest{0.0} & 11.1 & \tbest{55.2} & \second{21.4} & 28.9 & $+1.7\,[-3.9,\,7.2]$ \\
\midrule
\multicolumn{15}{c}{\cellcolor{black!7}\textit{Base model: Mistral-7B}}\\
Mem0 & 23.6 & 10.6 & 3.7 & \tbest{16.7} & 16.6 & $+4.0\,[1.8,\,6.2]$ & 34.5 & 26.7 & 0.0 & 3.7 & 24.1 & \tbest{32.1} & 20.2 & $+15.6\,[8.4,\,22.8]$ \\
A-Mem & 26.4 & 8.2 & 4.0 & 10.4 & 17.3 & $+3.3\,[0.8,\,5.3]$ & 17.2 & 20.0 & 0.0 & 0.0 & 6.9 & 10.7 & 9.2 & $+26.6\,[19.3,\,33.8]$ \\
Generative Agents & 5.2 & 4.6 & 2.8 & 7.3 & 4.7 & $+15.9\,[14.1,\,17.8]$ & 17.2 & 23.3 & 0.0 & 0.0 & 55.2 & 10.7 & 17.9 & $+17.9\,[11.3,\,25.1]$ \\
MemoryBank & 4.2 & 4.3 & 1.6 & 10.4 & 4.1 & $+16.5\,[14.2,\,19.4]$ & 37.9 & 30.0 & 0.0 & 3.7 & 10.3 & 10.7 & 15.6 & $+20.2\,[13.0,\,27.5]$ \\
SimpleMem & 3.9 & 2.8 & 0.9 & 4.2 & 3.1 & $+17.5\,[14.1,\,20.9]$ & 10.3 & 10.0 & 0.0 & 0.0 & 0.0 & 0.0 & 3.5 & $+32.3\,[25.6,\,39.0]$ \\
\cmidrule(l){1-15}
\rowcolor{blue!4} \method\ \textbf{(ours)} & \tbest{29.6} & \tbest{12.4} & \second{5.6} & \second{16.7} & \tbest{20.6} & --- & \second{51.7} & \second{50.0} & 0.0 & \tbest{22.2} & \tbest{58.6} & \second{32.1} & \tbest{35.8} & --- \\
\;\; consolidation-only & 5.1 & 6.7 & 3.1 & 13.5 & 5.5 & $+15.1\,[12.3,\,18.0]$ & 6.9 & 0.0 & \second{0.0} & 3.7 & 6.9 & 17.9 & 5.8 & $+30.1\,[23.4,\,37.3]$ \\
\;\; episodic-only & \second{28.9} & \second{12.1} & \tbest{5.6} & 15.6 & \second{20.1} & $+0.5\,[-1.5,\,2.4]$ & \tbest{58.6} & \tbest{50.0} & \tbest{0.0} & \second{7.4} & \second{55.2} & 17.9 & \second{31.8} & $+4.0\,[-1.0,\,9.0]$ \\
\bottomrule
\end{tabular}}
\caption{\textbf{Agentic memory under Cover-EM.} Identical layout to Table~\ref{tab:main},
scored with Cover-EM \cite{press2023compositionality,rosset2020knowledge} --- the normalised gold
answer appears in the prediction --- instead of the LLM judge. Unlike token-F1 this does not
penalise answer length, so systems with different generation styles remain comparable.}
\label{tab:mainCEM}
\end{table*}

\begin{table*}[t]\centering
\resizebox{\textwidth}{!}{%
\small\setlength{\tabcolsep}{4pt}\renewcommand{\arraystretch}{1.1}
\begin{tabular}{l c ccc >{\scriptsize}c >{\scriptsize}c ccc >{\scriptsize}c >{\scriptsize}c}
\toprule
& & \multicolumn{5}{c}{\textbf{LoCoMo}} & \multicolumn{5}{c}{\textbf{LongMemEval-S}} \\
\cmidrule(lr){3-7}\cmidrule(lr){8-12}
\textbf{System} & \textbf{rank} & \textsc{G} & \textsc{Q} & \textsc{M} & \normalsize $\Delta$ vs \method & \normalsize $\Delta$ vs consol.
 & \textsc{G} & \textsc{Q} & \textsc{M} & \normalsize $\Delta$ vs \method & \normalsize $\Delta$ vs consol. \\
\midrule
D2L-Merge & 8 per session & 1.1 & 1.7 & 0.0 & $+18.0\,[15.3,\,20.6]$ & $+5.7\,[4.3,\,7.2]$ & 0.6 & 2.3 & 3.5 & $+29.5\,[25.2,\,33.7]$ & $+5.4\,[2.9,\,7.9]$ \\
D2L-Retrieve & 8 & 5.0 & 4.4 & 6.1 & $+13.7\,[10.7,\,16.5]$ & $+1.5\,[-0.2,\,3.1]$ & 2.9 & 6.4 & 5.8 & $+26.6\,[22.5,\,30.8]$ & $+2.5\,[0.2,\,5.0]$ \\
Test-time LoRA\ttt & 8 & 0.6 & 5.0 & 2.8 & $+16.1\,[13.9,\,18.4]$ & $+3.9\,[2.2,\,5.8]$ & 2.3 & 2.3 & 0.0 & $+30.1\,[26.0,\,34.1]$ & $+6.0\,[3.7,\,8.3]$ \\
TLM\ttt & 8 & 0.6 & 3.9 & 2.2 & $+16.7\,[14.3,\,19.2]$ & $+4.4\,[3.0,\,6.0]$ & 1.7 & 3.5 & 3.5 & $+28.7\,[24.9,\,32.8]$ & $+4.6\,[2.5,\,6.9]$ \\
TMEM-once\ttt & 8 & 2.2 & 6.7 & 2.2 & $+15.2\,[12.7,\,17.8]$ & $+3.0\,[1.6,\,4.5]$ & 2.9 & 5.2 & 2.9 & $+27.9\,[24.1,\,32.0]$ & $+3.9\,[1.5,\,6.4]$ \\
TMEM-per-question\ttt & 8 & 6.1 & \second{17.8} & 10.0 & $+7.6\,[5.0,\,10.3]$ & $-4.6\,[-6.9,\,-2.3]$ & 2.3 & \second{29.5} & 4.0 & $+19.7\,[15.6,\,23.7]$ & $-4.4\,[-7.5,\,-1.3]$ \\
FT-Ceiling\ttt & full & \second{10.6} & 10.6 & \second{12.2} & $+7.8\,[5.1,\,10.7]$ & $-4.4\,[-6.5,\,-2.3]$ & 1.7 & 3.5 & 2.9 & $+28.9\,[24.9,\,32.9]$ & $+4.8\,[2.5,\,7.1]$ \\
\method\ consolidation-only & 8 & 7.2 & 6.7 & 6.1 & $+12.2$ & --- & \second{9.2} & 7.5 & \second{5.8} & $+24.1\,[20.0,\,28.3]$ & --- \\
\rowcolor{blue!7} \method\ \textbf{(ours)} & 8 & \tbest{20.0} & \tbest{20.0} & \tbest{16.7} & --- & & \tbest{28.3} & \tbest{30.6} & \tbest{35.8} & --- & \\
\bottomrule
\end{tabular}}
\caption{\textbf{Weight-space memory under Cover-EM.} Comparison shows delta over CLAM and consolidated channel only.}
\label{tab:paramCEM}
\end{table*}

\subsection{Full-Context Teacher}
\label{app:teacher}
\begin{figure}[t]
\centering
\includegraphics[width=0.8\columnwidth]{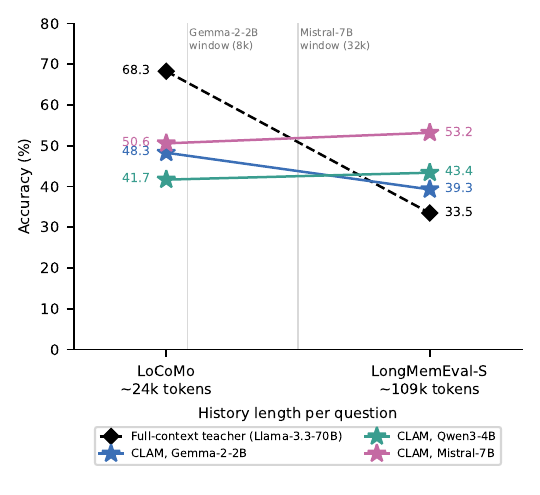}
\caption{\method\ vs.\ the full-context teacher as the history grows (same 180 questions per benchmark).}
\label{fig:context}
\end{figure}

Figure~\ref{fig:context} compares \method\ with the Llama-3.3-70B teacher, which answers with the
entire history in its context. On LoCoMo (${\approx}$24k tokens per history) the teacher scores 68.3;
on LongMemEval-S (${\approx}$109k tokens) it falls to 33.5, below \method\ on every base. Five of the 173
LongMemEval-S teacher calls failed and are scored as wrong, so 33.5 is a lower bound (at most 36.4).

\subsection{Decoding Variance}
\label{app:seeds}
Table~\ref{tab:seeds} repeats the LoCoMo evaluation with sampling (temperature 0.7, top-$p$ 0.95, three
seeds) on a 399-question stratified subset.

\begin{table}[h]
\centering
\resizebox{\columnwidth}{!}{%
\small
\begin{tabular}{lcc}
\toprule
\textbf{System} & \textbf{Greedy} & \textbf{Sampled, mean $\pm$ sd}\\
\midrule
\method, Gemma-2-2B   & 47.6 & 46.5 $\pm$ 1.04\\
\method, Qwen3-4B     & 49.1 & 49.7 $\pm$ 0.95\\
\method, Mistral-7B   & 50.6 & 50.5 $\pm$ 0.77\\
\bottomrule
\end{tabular}}
\caption{\textbf{Greedy vs.\ sampled decoding} (LoCoMo, 3 seeds).}
\label{tab:seeds}
\end{table}

\subsection{Episodic-Buffer Granularity}
\label{app:granularity}

The episodic buffer retrieves session-level units: each session, split on turn boundaries into pieces of at most 1{,}800 characters and carrying its session's date header. To test whether finer, sub-session episodes help, we vary only the buffer's unit and keep the consolidation channel, $k{=}5$, and the token budget fixed (Gemma-2-2B, LoCoMo, 180 questions). Coarser units work best (Table~\ref{tab:granularity}): finer pieces strip the surrounding turns that disambiguate references, and at a fixed $k$ they carry less evidence into the prompt.

\begin{table}[h]
\centering
\resizebox{\columnwidth}{!}{%
\small
\begin{tabular}{lcc}
\toprule
\textbf{Buffer unit} & \textbf{Units / conversation} & \textbf{Accuracy (\%)}\\
\midrule
Session pieces, $\le$1{,}800 chars (\method) & 68 & \textbf{45.6}\\
1{,}024-character chunks & 118 & 40.6\\
512-character chunks & 183 & 41.1\\
\bottomrule
\end{tabular}}
\caption{Episodic-buffer granularity (Gemma-2-2B, LoCoMo, $k{=}5$). Only the buffer's unit changes.}
\label{tab:granularity}
\end{table}

\subsection{Qualitative Examples}
Figures~\ref{fig:exFR}--\ref{fig:exAB} show one generated training trajectory for each of the six
memory types, drawn verbatim from the training corpus. Each panel gives the persona the history was
generated from, the number of sessions in that history, a few turns of the gold-evidence session and
one distractor session (two gold-evidence sessions for cross-session synthesis); the remaining sessions are elided.

\begin{figure}[t]\centering\small
\setlength{\fboxsep}{7pt}%
\fcolorbox{black!25}{black!2}{\begin{minipage}{0.93\columnwidth}
\textbf{Factual recall}\hfill{\scriptsize 14 sessions in history \textperiodcentered{} gold evidence in session 1}\\[2pt]
{\footnotesize\itshape The fact is stated once, in one session out of many; everything else in the history is topically similar noise, so the memory must keep the detail rather than the gist.}\\[5pt]
{\scriptsize\textbf{Persona}\; Nneoma Nkrumah, 33, Accra, Ghana -- Clinical Laboratory Technician (healthcare; expertise 2/5); terse casual register. Working on: implementing a new quality control system for blood sample analysis. Interests: has a pet hedgehog named Kuku; lives in a small compound house with her husband and two kids; enjoys marathon knitting and has made over 50 scarves...}\\[6pt]
{\scriptsize\textbf{2024-01-20 \textperiodcentered{} session 1}\,\textcolor{orange!85!black}{[gold evidence]}}\\[1pt]
{\scriptsize \textit{assistant:} ...a training program for lab technicians covers topics like laboratory safety, equipment operation, and sample handling... What's your current lab setup like? \\ \textit{user:} we use microscopes, centrifuges, and autoclaves \\ \textit{assistant:} Those are essentials. For a training program, you might want to create a hands-on module where trainees can practice using these machines under supervision...}\\[5pt]
{\scriptsize\textbf{2024-05-19 \textperiodcentered{} session 9}\,\textcolor{black!55}{[distractor]}}\\[1pt]
{\scriptsize \textit{user:} i'm trying to improve our lab's quality control \\ \textit{assistant:} Quality control is a critical aspect of laboratory work... Have you considered developing a training program for your lab staff to ensure they're up-to-date on the latest quality control procedures?}\\[6pt]
\hrule\vspace{5pt}
{\scriptsize\textbf{Q}\; List the equipment the user's lab uses}\\[2pt]
{\scriptsize\textbf{Gold}\; \textcolor{green!35!black}{Microscopes, centrifuges, and autoclaves}}
\end{minipage}}
\caption{Generated training trajectory: \textbf{factual recall} (\texttt{persona\_01276\_\_ckpt14\_\_q01}). One gold session and one distractor session are shown; the remaining sessions are elided.}
\label{fig:exFR}
\end{figure}

\begin{figure}[t]\centering\small
\setlength{\fboxsep}{7pt}%
\fcolorbox{black!25}{black!2}{\begin{minipage}{0.93\columnwidth}
\textbf{Preference inference}\hfill{\scriptsize 30 sessions in history \textperiodcentered{} gold evidence in sessions 12, 20}\\[2pt]
{\footnotesize\itshape The preference is never stated as a preference. It has to be inferred from how the user talks about their work, so verbatim retrieval of the session does not by itself yield the answer.}\\[5pt]
{\scriptsize\textbf{Persona}\; Kaiora Te Ariki, 32, Wellington, New Zealand -- Visual Effects Artist (creative; expertise 4/5); mixed register. Working on: creating visual effects for a NZ-based sci-fi film. Interests: has a rescued kakariki named Kiki; lives in a small apartment in the Aro Valley; is relearning te reo Maori; practices home fermentation...}\\[6pt]
{\scriptsize\textbf{2024-07-03 \textperiodcentered{} session 20}\,\textcolor{orange!85!black}{[gold evidence]}}\\[1pt]
{\scriptsize \textit{user:} ...I've been putting in some long hours, and I've had to fuel up on coffee. I used to always order a flat white at my favorite coffee shop, but lately, I've switched to long blacks. \\ \textit{assistant:} ...Long blacks can be quite strong, so I'm guessing you might need the extra caffeine kick to power through those long work sessions... \\ \textit{user:} Yeah, I think the long black gives me a bit more energy, which is helpful when I'm working on intricate details...}\\[5pt]
{\scriptsize\textbf{2024-03-26 \textperiodcentered{} session 9}\,\textcolor{black!55}{[distractor]}}\\[1pt]
{\scriptsize \textit{user:} ...I'm meeting with the director later to discuss the project... Maybe I'll grab a flat white at my favorite coffee shop beforehand to get my mind going \\ \textit{assistant:} A good cup of coffee can work wonders for creativity!...}\\[6pt]
\hrule\vspace{5pt}
{\scriptsize\textbf{Q}\; Is the user likely to prefer a coffee with a strong flavor?}\\[2pt]
{\scriptsize\textbf{Gold}\; \textcolor{green!35!black}{Yes, based on their switch from flat whites to long blacks}}
\end{minipage}}
\caption{Generated training trajectory: \textbf{preference inference} (\texttt{persona\_00553\_\_ckpt30\_\_q02}). One gold session and one distractor session are shown; the remaining sessions are elided.}
\label{fig:exPI}
\end{figure}

\begin{figure}[t]\centering\small
\setlength{\fboxsep}{7pt}%
\fcolorbox{black!25}{black!2}{\begin{minipage}{0.93\columnwidth}
\textbf{Cross-session synthesis}\hfill{\scriptsize 30 sessions in history \textperiodcentered{} gold evidence in sessions 12, 23, 28}\\[2pt]
{\footnotesize\itshape No single session contains the answer; it must be composed from evidence spread across sessions, which a top-$k$ retriever can only supply if it happens to return all the pieces at once.}\\[5pt]
{\scriptsize\textbf{Persona}\; Sof\'ia Elena G\'omez-Morales, 55, Buenos Aires, Argentina -- Abogada Especializada en Derecho Ambiental (legal; expertise 3/5); verbose casual register. Working on: Representing a local environmental group in a lawsuit against a polluting factory. Interests: Has a rescued Chacoan peccary named Rita; Enjoys foraging for wild mushrooms in the Paran\'a Delta...}\\[6pt]
{\scriptsize\textbf{2024-04-10 \textperiodcentered{} session 12}\,\textcolor{orange!85!black}{[gold evidence]}}\\[1pt]
{\scriptsize \textit{user:} I'll have to check that out. I'm actually also really into fungi - I've got a little collection of rare species that I've found on my travels. Do you know anything about mycology? \\ \textit{assistant:} Mycology is a fascinating field... \\ \textit{user:} That's so cool. I've been trying to learn more about the different types of fungi and how to identify them...}\\[5pt]
{\scriptsize\textbf{2024-08-16 \textperiodcentered{} session 28}\,\textcolor{orange!85!black}{[gold evidence]}}\\[1pt]
{\scriptsize \textit{user:} ...I just learned about this really cool fungus that can clean up pollutants in soil. Have you heard about it? \\ \textit{assistant:} That sounds fascinating! There are several species of fungi that have been found to have potential for bioremediation...}\\[6pt]
\hrule\vspace{5pt}
{\scriptsize\textbf{Q}\; Describe the user's experience with mycology.}\\[2pt]
{\scriptsize\textbf{Gold}\; \textcolor{green!35!black}{The user is interested in mycology and has been learning about different species of fungi, including their potential for bioremediation.}}
\end{minipage}}
\caption{Generated training trajectory: \textbf{cross-session synthesis} (\texttt{persona\_00427\_\_ckpt30\_\_q03}). Two of the three gold sessions are shown; the remaining sessions are elided.}
\label{fig:exCS}
\end{figure}

\begin{figure}[t]\centering\small
\setlength{\fboxsep}{7pt}%
\fcolorbox{black!25}{black!2}{\begin{minipage}{0.93\columnwidth}
\textbf{Knowledge update}\hfill{\scriptsize 18 sessions in history \textperiodcentered{} gold evidence in sessions 3, 18}\\[2pt]
{\footnotesize\itshape A later session contradicts an earlier fact. The gold answer is the \emph{current} value, so a system that surfaces the first mention is confidently wrong rather than merely uninformed.}\\[5pt]
{\scriptsize\textbf{Persona}\; Fatima Al-Khateeb, 30, Doha, Qatar -- Curriculum Developer for Secondary Education (education; expertise 4/5); mixed register. Working on: Developing a new curriculum for teaching Arabic language and literature to non-native speakers. Interests: Has a small apiary with 5 beehives in her backyard; Owns a sea kayak named `Dukhan'...}\\[6pt]
{\scriptsize\textbf{2024-10-01 \textperiodcentered{} session 18}\,\textcolor{orange!85!black}{[gold evidence]}}\\[1pt]
{\scriptsize \textit{user:} ...I've been thinking a lot about this and I have to admit, my views on standardized testing have changed recently. \\ \textit{assistant:} That's interesting! What made you change your mind about standardized testing? \\ \textit{user:} Well, I used to be completely against it, but I've been reading some research that suggests it can be useful in certain contexts, like measuring progress over time or comparing student performance across different schools.}\\[5pt]
{\scriptsize\textbf{2024-08-17 \textperiodcentered{} session 15}\,\textcolor{black!55}{[distractor]}}\\[1pt]
{\scriptsize \textit{user:} ...Do you have any suggestions on how to track student learning outcomes in a project-based learning environment? \\ \textit{assistant:} ...One approach is to use rubrics that outline specific criteria and standards for student projects. You can also use self-assessment and peer review...}\\[6pt]
\hrule\vspace{5pt}
{\scriptsize\textbf{Q}\; What is the user's current stance on standardized testing?}\\[2pt]
{\scriptsize\textbf{Gold}\; \textcolor{green!35!black}{The user now sees standardized testing as useful in certain contexts, such as measuring progress over time or comparing student performance across different schools; previously, they were completely against it.}}
\end{minipage}}
\caption{Generated training trajectory: \textbf{knowledge update} (\texttt{persona\_02278\_\_ckpt18\_\_q00}). One gold session and one distractor session are shown; the remaining sessions are elided.}
\label{fig:exKU}
\end{figure}

\begin{figure}[t]\centering\small
\setlength{\fboxsep}{7pt}%
\fcolorbox{black!25}{black!2}{\begin{minipage}{0.93\columnwidth}
\textbf{Temporal reasoning}\hfill{\scriptsize 14 sessions in history \textperiodcentered{} gold evidence in session 13}\\[2pt]
{\footnotesize\itshape The answer is a date or an ordering, which is carried by the session metadata and by the position of a session in the stream, not by its content.}\\[5pt]
{\scriptsize\textbf{Persona}\; Rukmini Deshpande, 21, Pune, India -- Research Assistant in the Department of Linguistics (academic; expertise 1/5); verbose formal register. Working on: Assisting a professor with a research project on the phonetics of Marathi dialects. Interests: Born and raised in a Marathi-speaking family in Pune; Has a pet parakeet named Sunanda...}\\[6pt]
{\scriptsize\textbf{2024-09-11 \textperiodcentered{} session 13}\,\textcolor{orange!85!black}{[gold evidence]}}\\[1pt]
{\scriptsize \textit{user:} That sounds really helpful. I'll definitely look into it. By the way, I've been trying to find some inspiration for my project and I recently attended a workshop on language pedagogy and linguistic theory. It was really enlightening and I got to meet some interesting people. \\ \textit{assistant:} Workshops can be a great way to learn from others and get new ideas...}\\[5pt]
{\scriptsize\textbf{2024-01-24 \textperiodcentered{} session 1}\,\textcolor{black!55}{[distractor]}}\\[1pt]
{\scriptsize \textit{user:} That's really helpful, thank you. I've recently attended a workshop on corpus linguistics and it's given me a lot of ideas for how to approach this project... \\ \textit{assistant:} Co-authoring a paper can be a great way to collaborate with others and share your research with a wider audience...}\\[6pt]
\hrule\vspace{5pt}
{\scriptsize\textbf{Q}\; When did the user attend a workshop on language pedagogy and linguistic theory?}\\[2pt]
{\scriptsize\textbf{Gold}\; \textcolor{green!35!black}{The user attended a workshop on language pedagogy and linguistic theory before September 11, 2024.}}
\end{minipage}}
\caption{Generated training trajectory: \textbf{temporal reasoning} (\texttt{persona\_02015\_\_ckpt14\_\_q00}). One gold session and one distractor session are shown; the remaining sessions are elided.}
\label{fig:exTR}
\end{figure}

\begin{figure}[t]\centering\small
\setlength{\fboxsep}{7pt}%
\fcolorbox{black!25}{black!2}{\begin{minipage}{0.93\columnwidth}
\textbf{Abstention}\hfill{\scriptsize 10 sessions in history \textperiodcentered{} no gold session (unanswerable)}\\[2pt]
{\footnotesize\itshape Nothing in the history answers the question. The gold answer is an explicit refusal, so a system that confabulates a plausible answer is penalised.}\\[5pt]
{\scriptsize\textbf{Persona}\; Kasper Vanderstappen, 48, Antwerp, Belgium -- Chief Operations Officer (business; expertise 5/5); terse technical register. Working on: Implementing a new supply chain management system across European operations. Interests: Has a collection of 300 antique door knockers from around the world; Competes in the annual Antwerp Bouldering League...}\\[6pt]
{\scriptsize\textbf{2024-06-15 \textperiodcentered{} session 6}\,\textcolor{black!55}{[nearest retrieved session---not evidence]}}\\[1pt]
{\scriptsize \textit{user:} Compliance is key. I've been listening to some Belgian jazz while working on my reports, trying to clear my head. Helps me focus. \\ \textit{user:} ...I've been putting in long hours, trying to get everything sorted. Just started playing the saxophone to unwind, but it's not as easy as it looks. \\ \textit{assistant:} ...The saxophone is a complex instrument, but with practice, you'll get the hang of it...}\\[5pt]
{\scriptsize\textbf{2024-04-22 \textperiodcentered{} session 4}\,\textcolor{black!55}{[distractor]}}\\[1pt]
{\scriptsize \textit{user:} Appreciate the help. Used to spend weekends foraging for wild mushrooms in the Belgian Ardennes, but had to cut back to one day a month due to increased work demands. Not much time for hobbies these days. \\ \textit{assistant:} It can be tough to balance work and personal life...}\\[6pt]
\hrule\vspace{5pt}
{\scriptsize\textbf{Q}\; What is the name of the user's saxophone teacher?}\\[2pt]
{\scriptsize\textbf{Gold}\; \textcolor{green!35!black}{I don't have any information about that from our conversations.}}
\end{minipage}}
\caption{Generated training trajectory: \textbf{abstention} (\texttt{persona\_00144\_\_ckpt10\_\_q11}). The nearest retrieved session and one distractor session are shown; the remaining sessions are elided.}
\label{fig:exAB}
\end{figure}

\end{document}